# CAFEEN: A Cooperative Approach for Energy Efficient NoCs with Multi-Agent Reinforcement Learning

Kamil Khan, Colorado State University; Sudeep Pasricha, Colorado State University

**Abstract:** In emerging high-performance Network-on-Chip (NoC) architectures, efficient power management is crucial to minimize energy consumption. We propose a novel framework called *CAFEEN* that employs both heuristic-based fine-grained and machine learning-based coarse-grained power-gating for energy-efficient NoCs. *CAFEEN* uses a fine-grained method to activate only essential NoC buffers during lower network loads. It switches to a coarse-grained method at peak loads to minimize compounding wake-up overhead using multi-agent reinforcement learning. Results show that *CAFEEN* adaptively balances power-efficiency with performance, reducing total energy by 2.60× for single application workloads and 4.37× for multi-application workloads, compared to state-of-the-art NoC power-gating frameworks.

———————— ◆ ————————

## 1 INTRODUCTION

The number of computing cores in Systems-on-Chip (SoCs) has drastically increased to meet the ever-growing demand for higher performance in emerging applications. Network-on-Chip (NoC) architectures are scalable, predictable, and programmable fabrics for meeting the communication needs of on-chip cores. Modern NoCs can efficiently connect hundreds to thousands of cores in various topologies, with mesh being a popular choice due to its simplicity.

Such NoCs are generally designed to accommodate peak load scenarios, using virtual channels (VCs) with several buffers per router input port for efficient channel utilization and Quality of Service (QoS). However, operating load is often significantly lower than the peak-load [1], necessitating efficient NoC power-gating (PG). The PG approach can selectively power down idle NoC components, using fine-grained or coarse-grained methods. Fine-grained PG targets specific NoC router components such as VCs, offering increased efficiency at the cost of complexity and potential peak load performance constraints. Coarse-grained PG in contrast, simplifies control by powering down entire NoC routers [1], [2], [3], [4]. While more effective under uniform router component utilization, coarse-grained PG can cause unnecessary leakage power consumption in idle components when utilization is uneven. Moreover, coarse-grained PG can result in high packet latency due to the contribution of wake-up latencies when multiple power-gated NoC routers are encountered by a packet along its path. Finally, the presence of PG routers complicates NoC routing.

In this article, we propose *CAFEEN*, a novel PG framework for NoCs that adaptively transitions between fine-grained and coarse-grained PG based on traffic load conditions. *CAFEEN* employs fine-grained PG to activate only necessary input buffers during low-load conditions. To optimize performance at higher traffic volumes, *CAFEEN* introduces a multi-agent reinforcement learning (MARL)-based routing framework to manage coarse-grained PG. When the NoC transitions to coarse-grained PG under the MARL-based framework, routing agents adaptively route NoC packets based on real-time traffic and network power state, optimizing power efficiency and performance through cooperation between multiple agents. The novel contributions of our CAFEEN framework include:

- We identify and quantify significant opportunities for power savings with fine-grained PG to enhance course-grained PG methods in NoCs;
- We develop a novel fine-grained PG strategy for low traffic load conditions in NoCs;
- For managing coarse-grained PG under high traffic load, we formulate a multi-agent reinforcement learning (MARL)-based framework using cooperative routing agents;
- We compare our framework, *CAFEEN* against state-of-the-art routing and PG methods for NoCs.

## 2 BACKGROUND

**Power-gating (PG) for Mesh NoCs with XY Routing**

Power-gating (PG) reduces static power consumption in NoCs by selectively powering down idle components. PG can be applied at various granularities, such as fine-grained PG targeting specific router components (e.g., input/output buffers, virtual channels, crossbar) or coarse-grained PG that powers down entire routers. A wake-up event triggers the router to power-up the required resources (fine-grained) or the entire router (course-grained). This power-up phase has its own associated power and performance overhead.

In 2D mesh NoCs with dimension-order routing (e.g., XY routing), PG can exploit the prevalence of "straight" packets. Straight packets traverse the network without requiring a 90-degree turn, and thus do not need to use the route computation and switch allocation stages. Despite requiring minimal functionality, due to their prevalence in XY routing, straight packets result in the highest number of router wake-ups. To address this issue, the "Turn-on-on-Turn" (TooT) approach [1] proposes the use of a low-power bypass link for straight packets, allowing the router to remain power-gated until a turning packet is encountered. The TooT bypass is implemented using a single forwarding buffer per router to store straight packets and a TooT controller to check if a packet needs to turn and control the bypass link accordingly. By default, TooT employs a coarse-grained PG policy, which powers up the entire router when a turning packet is encountered on any of the input ports.

**Multi-Agent Reinforcement Learning**

Multi-agent reinforcement learning (MARL) extends reinforcement learning (RL) to scenarios involving multiple agents interacting within a shared environment. In RL, an agent observes the state of its environment $s_t$, takes an action $a_t$, and receives feedback as a reward $r_t$, with the objective of maximizing cumulative rewards over a series of actions. This approach suits situations lacking explicit instructions, where agents learn through exploration. The behavior of an RL agent is determined by a policy that maps states to actions. A popular RL algorithm for learning optimal policies is Q-learning [5], with its update equation given by:

$$Q_{new}(s_t, a_t) \leftarrow (1-\alpha)Q(s_t, a_t) + \alpha(r + \gamma \max_{a \in A} Q(s_{t+1}, a)) \quad (1)$$

Where:
- $Q_{new}(s_t, a_t)$ is the updated Q-value for taking action $a_t$ at state $s_t$, at time $t$
- $Q(s_t, a_t)$ is the current Q-value for taking action $a_t$ at state $s_t$, at time $t$
- $\alpha$ is the learning rate, $0 \leq \alpha \leq 1$
- $r$ is the scalar reward received after taking action $a_t$ in state $s_t$
- $\gamma$ is the discount factor, $0 \leq \gamma \leq 1$
- $\max_{a \in A} Q(s_{t+1}, a)$ is the maximum Q-value for the next state $s_{t+1}$ over all actions $a$ in set $A$

In MARL, multiple RL agents interact with a shared environment. Each agent's actions can influence the environment, necessitating communication of goals and rewards among all agents. Cooperative MARL scenarios have agents working together towards a shared goal, aligning their actions to maximize a common reward. In PG-enabled NoCs, we utilize cooperative RL agents to adapt routing decisions to a dynamic traffic and power environment, optimizing network performance and PG efficiency by minimizing unnecessary router wake-ups.

## 3 CAFEEN FRAMEWORK OVERVIEW

In this section, we describe *CAFEEN*, our proposed Cooperative and Adaptive Framework for Energy-Efficient NoCs, which overcomes the drawbacks of prior work. First, we discuss our proposed fine-grained power gating approach and its limitations under high traffic load. Then, we introduce our collaborative MARL policy for efficient coarse-grained PG during high traffic load.

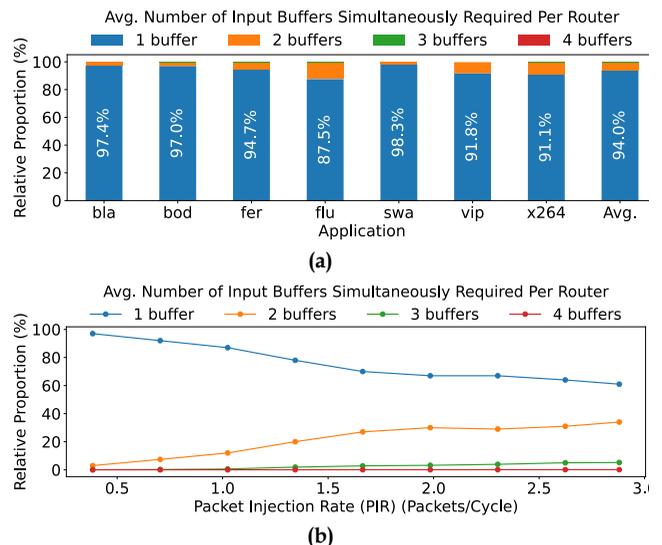

Figure 1. (a) Most TooT routers turn a single packet at a time, requiring only a single input buffer in the NoC router to be activated. (b) The proportion of packets which require multiple input buffers per router to be activated increases with increasing PIR for uniform random traffic.

**Enhancing TooT-based NoCs with Fine-grained PG**

The baseline TooT mechanism [1] employs coarse-grained PG. We identify a significant opportunity for fine-grained PG in TooT-based NoCs, particularly under low-load conditions prevalent in SoC platforms. Our analysis of PARSEC applications using TooT with XY routing reveals that 94% of the time, only one out of four input buffers is required by turning packets (See Fig. 1(a)). Consequently, most input buffers under the baseline coarse-grained TooT approach [1] are not utilized.

We propose introducing fine-grained PG in the TooT approach (Fig. 2), building on the observation that multiple input buffers are rarely needed under typical low-load conditions. Our approach manages individual input buffers separately to reduce leakage power, ensuring that only the input buffers catering to turning packets are selectively powered on, while those not in use remain in a power-gated state. We use a threshold of idle cycles $t_{idle}$ to initiate PG for the input buffers, chosen by considering the break-even time (BET) — the duration a component must remain inactive to balance the energy costs of power mode transition.

However, with an increase in NoC traffic load, the likelihood of encountering multiple turning packets simultaneously increases. Fig. 1(b) shows that at high-loads in a mesh NoC, the incidence of simultaneous turning packets — and consequently, the need for multiple buffer wake-ups — increases. At high traffic load, waking up multiple input buffers separately for each turning packet introduces cumulative wake-up latencies in our proposed fine-grained PG approach, which can deteriorate performance compared to coarse-grained PG.

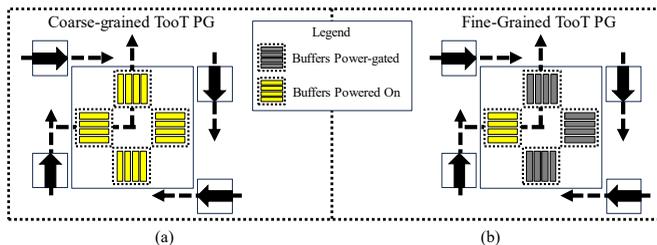

**Figure 2. (a)** Traditional coarse-grained PG, including TooT powers on the entire router including all input buffers **(b)** Proposed fine-grained PG powers on only the required input buffer.

## Coarse-grained PG using Multi Agent Reinforcement Learning (MARL)

Under high-load conditions, we introduce the ability for individual NoC routers to adaptively transition to coarse-grained power gating (PG) mode. Coarse-grained PG enables simultaneous activation of all router resources, effectively masking the multiple wake-up latencies that would occur for multiple packet arrivals in fine-grained PG. Each router independently transitions between fine-grained and coarse-grained PG modes based on traffic conditions. In coarse-grained PG mode, NoC routers form a collaborative MARL agent network. The MARL approach aims to maximize energy savings by minimizing unnecessary router wake-ups through intelligent routing. Each router hosts an RL agent responsible for determining routing paths for packets injected by the processing element (PE).

### MARL Problem Formulation

In our MARL framework, each RL agent is located in a router and is responsible for determining the routing path for all packets injected by the PE (see Fig. 3(a)). The state $s_t$ is defined as the destination of the packet. For every state, an action $a_t$ defines a routing path. To simplify the problem and reduce the state-action space, we limit the routing actions to either the XY or YX path for the packet to reach its destination. This choice is motivated by two key factors:

- Power efficiency: When using XY or YX routing, only a single router needs to be powered on (to turn the packet) when the source and destination are in different columns or rows. This minimizes the number of router wake-ups compared to fully adaptive routing.
- Problem complexity reduction: By limiting the choices to XY or YX, we significantly reduce the total number of possible paths between all pairs of nodes. This accelerates the learning process and decreases the agent's footprint.

The primary optimization goal of our MARL framework is to minimize the number of router wake-ups, which incur significant power and performance costs in NoC power-gating. Equivalently, we aim to maximize the efficient utilization of already awake routers. To achieve this in a multi-agent setting, we design a reward function that encourages cooperation among agents to optimize NoC power-efficiency through optimal routing actions.

We define the reward $r_t$ as the total number of packets that turned during a reward epoch, representing power efficiency. It encourages agents to reuse existing waking routers instead of waking up sleeping routers. A reward epoch is triggered when a packet reaches its turning router, as shown in Fig. 3(b). The epoch continues for a fixed duration of $t_{epoch}$ cycles or until the router re-enters power-gating, whichever occurs first.

The sum of packets turned during a reward epoch serves as a shared reward, encouraging cooperative behavior among routing agents. This reward structure motivates agents to favor routes with already active routers, minimizing unnecessary wake-ups and maximizing the reuse of powered-on routers. Consequently, agents are driven to maximize packet turns within an ongoing epoch, reducing the need for new wake-up cycles and allowing idle routers to remain power-gated for longer durations.

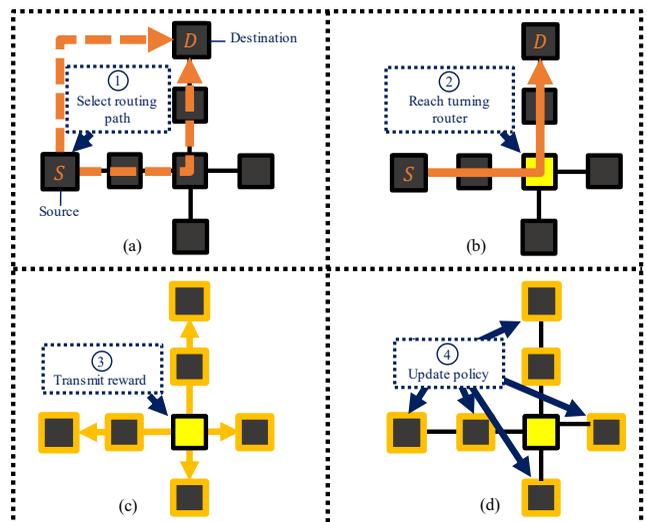

**Figure 3.** The update mechanism for a single agent for handling the flow S→D: (a) routing agent at the source router selects an action, (b) a reward epoch starts, (c) reward is broadcast, (d) the agents update their policies.

*Q-Learning Implementation*

We implement our MARL framework using Q-learning, allowing agents to learn optimal policies through experience. Each router maintains a Q-value table representing the expected power-efficiency of routing paths. To minimize memory requirements, we use destination row and column as state indicators, rather than individual node IDs [6] (Table 1). This simplification is possible because in deterministic-order routing on a mesh topology, all packets from a given row of source nodes to a specific column of destination nodes (and vice versa) share the same turning router.

The update mechanism for our Q-learning implementation is based on a simplified version of the standard Q-learning update equation. We use Equation 2, a reduction of Equation 1, to update the Q-values:

$$Q_{new}(s,a) \leftarrow (1-\alpha)Q(s,a) + \alpha r \qquad (2)$$

This reduction is possible because our problem is structured as a single state-action step to the terminal state. By using only XY and YX as possible paths, we eliminate the need $\gamma \max_{a \in A} Q(s_{t+1}, a)$ from the standard equation, as there are no future states to consider after the routing decision is made. The agents' collective goal is to determine the most power-efficient path (XY or YX) for each source-destination pair.

The update process is triggered when a router is powered on by a turning packet in coarse-grain PG mode. After accumulating the reward over the epoch, the router broadcasts a single-flit integer reward over a dedicated channel to each agent in the same row and column. Since it always travels in a straight line, the reward flit does not require turning. The dedicated channel and straight path allows for a simple flit structure with just the integer reward value. Upon receiving the reward flit, each agent updates the Q-value for the state and action corresponding to the turning router according to Equation 2. For instance, as illustrated in Fig. 3(c), all routers in the same row as the turning router receiving the reward flit will update the value $Q(XY)$ for the column state of the turning router. Similarly, routers in the same column will update $Q(YX)$ for the row state of the turning router. We note that the computational overhead for MARL is the same as any table-based routing algorithm, requiring a table look-up operation for determining an entire routing path. Since the update operation is not in the critical path of routing, it does not incur additional latency.

To balance exploration and exploitation, we employ an ε-greedy policy, typically selecting the action with the higher Q-value but allowing random actions with a small probability ε.

This Q-learning implementation enables routers to cooperatively learn and adapt their routing policies, collectively minimizing unnecessary wake-ups and maximizing active router reuse. With sufficient exploration, Q-values converge to a solution optimally balancing power efficiency and network performance.

*Cooperative Routing Policy*

CAFEEN's MARL framework enables a cooperative routing policy that adapts to varying network conditions and ensures deadlock-free operation. Agents use shared rewards to update Q-values, minimizing unnecessary wake-ups and maximizing the reuse of active routers. The policy dynamically switches between fine-grained PG (low traffic) with deadlock-free XY routing and coarse-grained PG (high traffic) using a learned XY-YX routing strategy based on current Q-values.

To prevent deadlocks in adaptive routing, we employ virtual channel (VC) partitioning, splitting VCs into sets restricted to either south-first or north-first turns, thereby eliminating cyclic dependencies while preserving routing flexibility.

**Table 1. Router Table for Storing Q-values (R×C Mesh)**

| Destination | $Q(XY)$ | $Q(YX)$ |
|---|---|---|
| $Row1$ | $Q(Row1, XY)$ | $Q(Row1, YX)$ |
| $Col1$ | $Q(Col1, XY)$ | $Q(Col1, YX)$ |
| … | … | … |
| $RowR$ | $Q(RowR, XY)$ | $Q(RowR, YX)$ |
| $RowC$ | $Q(RowC, XY)$ | $Q(RowC, YX)$ |

## 4 EXPERIMENTAL RESULTS

**Experimental Setup**

We compare our *CAFEEN* framework against several state-of-the-art frameworks: *No PG* (baseline NoC without power gating, using XY routing), *Conv+XY* (conventional power gating where routers sleep when idle and wake up on injection or packet traversal, using XY routing), *TooT* [1] (TooT bypass for straight packets, XY routing), *SMART* [2] (deterministic XY/YX paths between node pairs), *SPONGE* [3] (always-on central column, non-minimal routing), and *Flov* [4] (always-on eastmost column, non-minimal adaptive routing). All bypass-enabled frameworks use coarse-grained PG (powering down entire routers), while *CAFEEN* employs both fine-grained (powering down individual router components) and coarse-grained PG, enhanced with multi-agent reinforcement learning.

We used the NoC simulator Noxim [7] to evaluate the performance of all frameworks. For traffic, we consider both real applications from the PARSEC benchmark suite and synthetic workloads. All synthetic workloads are run until 1 million packets are drained. We use Netrace [8] to preserve NoC packet dependency relationships in PARSEC applications using gem5 instrumentation. We implemented the hardware for our *CAFEEN* framework and all prior works in RTL. All designs were then synthesized using Cadence Genus to obtain power and area estimates. For estimating area of the baseline NoC router, we used ORION 3.0 [9].

We consider a 64-core SoC connected via an 8×8 input-buffered NoC with 2D-mesh topology, implemented in 45 nm technology (1V, 1 GHz) and scaled to 14 nm using DeepScaleTool [10]. Each NoC node has a PE and a router with 5 input/output ports (4 VCs per port, 4 flit buffers per VC, 128-bit flits). Our router includes a bypass latch (1 flit wide per VC) for storing and traversing straight packets during power gating. The Q-value table has 16 rows (one per row/column ID) and 2 columns (XY/YX paths). For fine-grained buffer PG, $t_{idle}$ and $t_{on}$ are set to 2 cycles; for coarse-grained TooT, they are 4 and 8 cycles [1]. The length of the reward epoch for MARL, $t_{epoch}$ is set to 16 cycles, the learning rate, $\alpha$ is set to 0.01, and the exploration rate, $\epsilon$ is set to 0.05. The simulator uses accurate energy estimates from the synthesized design for the bypass, power management, adaptive routing, and Q-value table.

### Results for PARSEC Traffic

Fig. 4(a) shows that CAFEEN outperforms other policies in normalized total NoC energy consumption for PARSEC single application workloads due to fine-grained PG. *CAFEEN* leads to 2.6× total energy reduction compared to *SMART* which performs the next best. This confirms the effectiveness of integrating fine-grained PG to enhance TooT and the novel MARL routing approach to improve power efficiency. *SPONGE* performs worse because of the non-minimal routing algorithm operating under low load. Finally, *Flov* performs the worst because of a non-minimal routing policy, as well as the idleness detection logic based on PE activity as opposed to NoC router activity. It was also observed that *Flov* routers could get stuck in the draining state, causing deadlock if any two routers drained through each other.

Fig. 4(b) shows the energy results for multi-application PARSEC benchmarks generated by simultaneously executing multiple PARSEC applications and supporting their higher-load traffic on the NoC. In this scenario, *CAFEEN* benefits from the coarse-grained PG mode using MARL to maximize power-efficiency. *CAFEEN* leads to 4.37× total energy reduction compared to *TooT* which performs the next best. Frameworks that make use of non-minimal routing policies, such as *Flov* and *SPONGE* suffer from greater congestion on the always-on routers due to increased load.

Fig. 5(a) and Fig. 5(b) show the average network packet latency overhead of *CAFEEN* compared to other frameworks for the single and multiple PARSEC application workload scenarios, respectively. *CAFEEN* adds a minimal 5.9% and 7.1% latency overhead on average compared to the baseline *No PG* approach, for single and multiple application workloads respectively. Compared to all PG frameworks, *CAFEEN* has the lowest latency overhead, highlighting the minimal performance impact of our approach while aggressively minimizing NoC energy consumption.

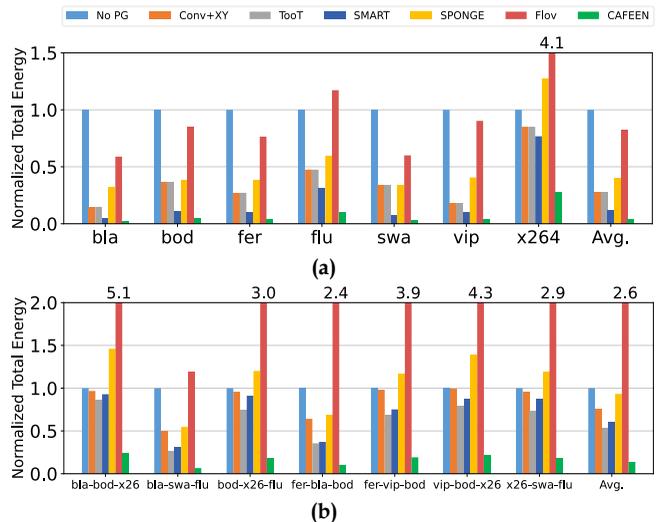

Figure 4. Normalized total energy for PARSEC (a) single application workloads; (b) multiple application workloads (combinations of three workloads executing simultaneously).

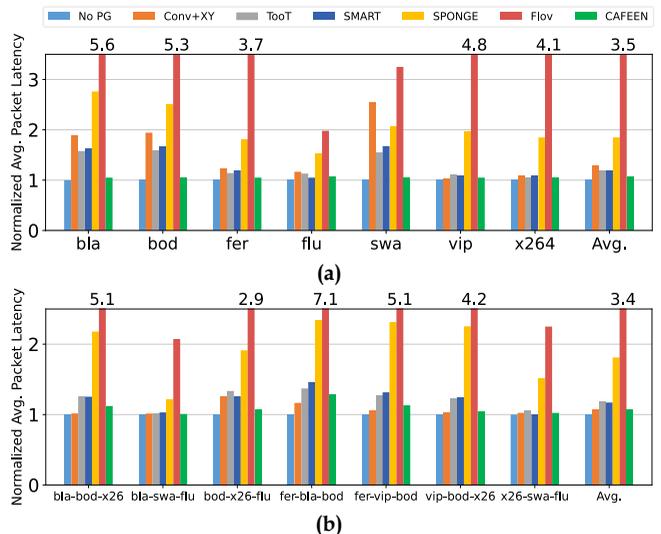

Figure 5. Normalized average packet latency for PARSEC (a) single application workloads; (b) multiple application workloads (combinations of three workloads executing simultaneously).

### Results for Synthetic Traffic

Fig. 6(a) shows that *CAFEEN* achieves the best energy efficiency across the bitreversal, transpose, and shuffle synthetic traffic patterns at low and high traffic loads. The only exception is uniform random traffic at high injection rates because of the lack of a learnable pattern in random traffic. However, scenarios with uniform random traffic are unlikely to be encountered in real application scenarios. Deterministic routing algorithms such as XY optimally distribute uniform random traffic across the network and will always perform better than any adaptive routing policy. *CAFEEN* performs the best for uniform random traffic out of all adaptive policies.

Fig. 6(b) shows the normalized execution time for the same workload. At low injection rates, the execution time is dominated by idle cycles, with negligible variation. However, as PIR increases, the execution time is impacted by congestion and router wake-up overhead. In all cases except uniform random traffic, our MARL approach in *CAFEEN* finds the best routing solution by using the same turning routers to avoid wake-up delays.

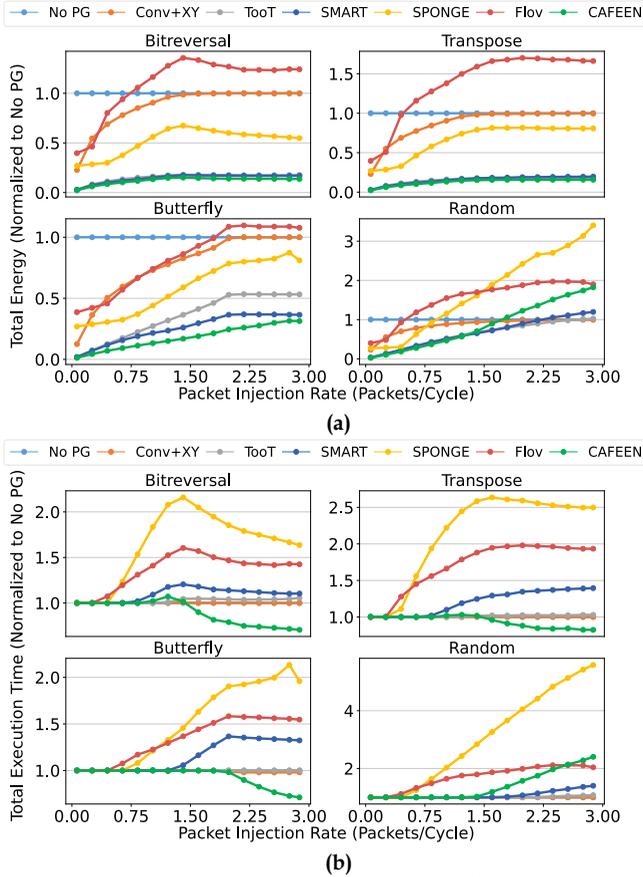

**Figure 6.** (a) Normalized total energy, (b) Normalized total execution time for increasing PIR of synthetic traffic patterns.

### Area Results

Lastly, we present the area overhead analysis of *CAFEEN*. Each router in *CAFEEN* can execute the multi-agent RL algorithm with a 4-bit 2D Q-table of 16 states and 2 actions per state, implemented via register arrays. Key optimizations include power-gating the table of Q-values except under high-load/coarse-grained PG, and using row/column IDs as states, resulting in linear state space growth (16 states for an 8×8 NoC instead of 64). The power state management, router table, and bypass add 2.62 µm², 74.60 µm², and 62.64 µm² area overhead, respectively. In total, *CAFEEN* adds 4.32% area overhead to the baseline NoC router, which is 2.57× less than conventional Q-routing [6], and adds 1.99% and 1.78% area overhead compared to state-of-the-art TooT-based NoC routers and Flov respectively.

## 5 CONCLUSIONS

NoC power consumption has become a significant portion of the power budget in emerging SoC platforms. In this work we introduce fine-grained power gating (PG) in TooT-based NoCs to significantly reduce static power consumption. We also use of cooperative MARL for coarse-grained PG management under high traffic load scenarios. Our *CAFEEN* framework reduces NoC energy consumption by 2.60× for single application workloads and 4.37× for multi-application workloads compared to the best state-of-the-art NoC PG frameworks from prior work. *CAFEEN* achieves these improvements while having a minimal impact on network latency and area. *CAFEEN* thus represents a promising framework to realize energy-efficient NoCs in emerging manycore SoC platforms.

**Kamil Khan** (kamil@colostate.edu) is a current Ph.D. student at Colorado State University. His research interests include reinforcement learning, many-core resource management, network-on-chips, and memory.

**Sudeep Pasricha** (sudeep@colostate.edu) received his Ph.D. in computer science from the University of California, Irvine in 2008. He is currently a Professor at Colorado State University. His research interests include networks-on-chip, and hardware/software co-design for energy-efficient, secure, and fault-tolerant embedded systems. He is an IEEE Fellow.